\documentclass[nohyperref]{article}

\usepackage{microtype}
\usepackage{graphicx}
\usepackage{booktabs} 

\usepackage{hyperref}

 \usepackage[accepted]{icml2022}

\usepackage{amsmath}
\usepackage{amssymb}
\usepackage{mathtools}
\usepackage{amsthm}
\usepackage{subfig}

\usepackage[capitalize,noabbrev]{cleveref}

\theoremstyle{plain}

\theoremstyle{definition}

\theoremstyle{remark}

\usepackage[textsize=tiny]{todonotes}

\icmltitlerunning{Multi-step domain adaptation by adversarial attack to $\mathcal{H} \Delta \mathcal{H}$-divergence}

\begin{document}

\twocolumn[
\icmltitle{Multi-step domain adaptation by adversarial attack to $\mathcal{H} \Delta \mathcal{H}$-divergence}



\begin{icmlauthorlist}
\icmlauthor{Arip Asadulaev}{Itmo,Airi}
\icmlauthor{Alexander Panfilov}{Tubingen}
\icmlauthor{Andrey Filchenkov}{Itmo}
\end{icmlauthorlist}

\icmlaffiliation{Itmo}{ITMO University, Saint-Petersburg, Russia}
\icmlaffiliation{Airi}{Artificial Intelligence Research Institute, Moscow, Russia}
\icmlaffiliation{Tubingen}{Tubingen Universitat, Germany}

\icmlcorrespondingauthor{Arip Asadulaev}{aripasadulaev@itmo.ru}

\icmlkeywords{Machine Learning, ICML}

\vskip 0.3in
]



\printAffiliationsAndNotice{} 

\begin{abstract}
Adversarial examples are transferable between different models. In our paper, we propose to use this property for multi-step domain adaptation. In unsupervised domain adaptation settings, we demonstrate that replacing the source domain with adversarial examples to $\mathcal{H} \Delta \mathcal{H}$-divergence can improve source classifier accuracy on the target domain. Our method can be connected to most domain adaptation techniques. We conducted a range of experiments and achieved improvement in accuracy on Digits and Office-Home datasets. 
\end{abstract}\vspace{-4mm}
\vspace{-5mm}\section{Introduction}
In domain adaptation, we train a model to make correct predictions on the different domains named source and target~\cite{DavidBCKPV10}. Domain adaptation techniques aim to bring the target domain ``closer‘’ to the source domain or find a common representation for these domains.~\cite{GaninL15, cdan}. 

Theoretically, the accuracy of domain adaptation is bounded by the source domain error and the divergence $\mathcal{H} \Delta \mathcal{H}$ between source and target domains~\cite{DavidBCKPV10,DavidLLP10,GermainHLM13}.  The divergence $\mathcal{H} \Delta \mathcal{H}$ can be approximated by the binary classifier between source and target domains. We find that source class samples turned into the target class sample (by an adversarial attack to the $\mathcal{H} \Delta \mathcal{H}$ neural approximation) are actually "closer" to the real target samples. We show that training the source classifier on these samples can improve its accuracy in the target domain. 

We call our method $\mathcal{H} \Delta \mathcal{H}$-divergence Domain Adaptation (HDA). We tested a range of domain adaptation algorithms coupled with the proposed technique on datasets ``Digits’’~\cite{mnisthandwrittendigit} and ``Office-Home’’~\cite{officehome}. 

\section{Background}
\textbf{Adversarial Attacks:} Having  sample $x$, target label $y$, model parameters $\theta$, loss function $L$, we can apply adversarial attack~\cite{DBLP:journals/corr/SzegedyZSBEGF13,GoodfellowSS14,Moosavi-Dezfooli16,MadryMSTV18,LuoLWX18}. For example, Fast Sign Gradient Descent (FSGD)~\cite{GoodfellowSS14} can be presented  as: ${x}_{i+1}={x}_{i}+\varepsilon \operatorname{sign}\left(-\nabla_{x} L\left(\theta, {x}_{i}, y_{i}\right)\right)$, with perturbation size $\varepsilon$ around the original image $x_i$. It was shown that  adversarial examples are transferable between different models~\cite{Transfer}. Recent studies have revealed that the adversarial examples are largely invariant to the models trained on different sampled datasets~\cite{DBLP:journals/corr/PapernotMG16}. 

\begin{table*}[h!]
\centering
\vskip 0.1in
\begin{center}
\begin{footnotesize}
\begin{sc}

\begin{tabular}{lllllll} 
\hline

Model       & MNIST$\rightarrow$MNIST-M & MNIST-M$\rightarrow$MNIST      & SVHN$\rightarrow$MNIST         & MNIST$\rightarrow$SVHN        &USPS$\rightarrow$MNIST \\ \hline

DANN        & 97.2 +- 1.37          & \textbf{73.4 +- 1.71}                  & 21.3 +- 2.38                  & 67.3 +- 3.20                           & 97.2 +- 1.37            \\ 
CDAN        & 97.8 +- 0.22          & 61.0 +- 2.76                    & 15.7 +- 4.55                           & 61.5 +- 5.14                          & 97.8 +- 0.22            \\ 
CDAN-E      & 97.8 +- 0.41          & 67.0 +- 4.32                    & 13.9 +- 1.47                           & 60.5 +- 4.44                         & 97.8 +- 0.41             \\ 
DAN         & 97.6 +- 1.17          & 43.6 +- 1.93                    & 19.2 +- 2.97                           & 61.2 +- 2.27                         & 97.6 +- 1.17           \\ 
JAN         & 97.6 +- 0.39          & 44.7 +- 4.48                   & 10.7 +- 1.70                            & 59.9 +- 3.14                     & 97.6 +- 0.39         \\ 
SHOT        & 98.1 +- 1.22          & 77.4 +- 1.44                   & 24.4 +- 1.50                          & 98.9 +- 2.11                         &98.0 +- 0.51           \\ \hline

DANN + HDA  & \textbf{98.1 +- 0.25} & 70.7 +- 2.86                  & \textbf{24.5 +- 3.47}                  & \textbf{78.6 +- 7.99}                  & \textbf{98.1 +- 0.25}            \\ 
CDAN + HDA  & \textbf{98.0 +- 0.41} & \textbf{65.2 +- 3.69}         & \textbf{17.2 +- 2.24}                  & \textbf{70.8 +- 8.29}                 & \textbf{98.0 +- 0.41}            \\ 
CDAN-E + HDA & \textbf{98.2 +- 0.03} & \textbf{69.9 +- 2.28}         & \textbf{16.0 +- 1.46}                  & \textbf{72.8 +- 3.95}                  & \textbf{98.2 +- 0.03}           \\ 
DAN + HDA   & \textbf{97.6 +- 0.15}  & \textbf{44.6 +- 1.24}         & \textbf{23.2 +- 2.91}                  & 60.7 +- 3.34                  & 97.6 +- 0.15            \\ 
JAN + HDA   & 97.1 +- 0.16          & \textbf{49.8 +- 6.47}         & \textbf{13.6 +- 2.18}                  & 59.3 +- 10.56                  & 97.1 +- 0.16            \\ 
SHOT + HDA  & \textbf{98.2 +- 0.45}          &  \textbf{77.6 +- 1.91}                            & \textbf{42.6 +- 3.21}                  & 98.9 +- 1.72                & 98.0 +- 0.40            \\ \hline
\end{tabular}
\end{sc}
\end{footnotesize}
\end{center}

\caption{Results of domain adaptation with HDA on Digits datasets.}
\label{tab:table-1}
\end{table*}
\begin{table*}[h!]
\centering
\begin{center}
\begin{footnotesize}
\begin{sc}
\begin{tabular}{llllllllllllll}
\hline
Model           & A$\rightarrow$C & A$\rightarrow$P & A$\rightarrow$R & C$\rightarrow$A & C$\rightarrow$P & C$\rightarrow$R 
                & P$\rightarrow$A & P$\rightarrow$C & P$\rightarrow$R & R$\rightarrow$A & R$\rightarrow$C & R$\rightarrow$P \\ \hline
DANN            & 45.6 & 59.3 & 70.1 & 47.0 & 58.5 & 60.9 & 46.1 & 43.7 & 68.5 & 63.2 & 51.8 & 76.8 \\ 
DAN            & 43.6 & 57.0 & 67.9 & 45.8 & 56.5 & 60.4 & 44.0 & 43.6 & 67.7 & 63.1 & 51.5 & 74.3\\ 
CDAN-E           & 50.7 & 70.6 & 76.0 & 57.6 & 70.0 & 70.0 & 57.4 & 50.9 & 77.3 & 70.9 & 56.7 & 81.6  \\ 
SHOT            & 57.1 & 78.1 & 81.5 & 68.0 & 78.2 & 78.1 & 67.4 & 54.9 & 82.2 & 73.3 & 58.8 & 84.3 \\ \hline
SHOT+HDA       & 57.0 & \textbf{78.5} & 80.6 & \textbf{69.1} & \textbf{79.5} & \textbf{78.4} &\textbf{68.3} & \textbf{55.5} & 81.5 & \textbf{73.7} & \textbf{60.5} & 83.8 \\ \hline
\end{tabular}
\end{sc}
\end{footnotesize}
\end{center}
\caption{Results on Office-Home datasets. Domains are: Art (A), Clipart (C), Product (P), Real-World (R)}
\label{tab:table-2}
\end{table*}

\textbf{Domain Adaptation:} For the given two domains, source $S_X$ and target $T_X$ over the input space $X$, the goal is to bring the target domain or its representation ``closer‘’ to the source domain. Following the learning theory~\cite{KiferBG04, DavidBCKPV10,DavidLLP10}, the domain adaptation error is bounded by the source domain classifier error and the $\mathcal{H} \Delta \mathcal{H}$ discrepancy between source and target domains. It is impossible to estimate exactly the value of $\mathcal{H} \Delta \mathcal{H}$, but it can be approximated by a binary classifier that discriminates the source and the target samples~\cite{DavidBCKPV10,DavidLLP10}. 

\section{Algorithm}

First of all, we train a binary classifier $\mathcal{H}_\omega$ between domains $S_X$ and $T_X$ that approximates $\mathcal{H} \Delta \mathcal{H}$. Then, we apply perturbations in the input space of $\mathcal{H}_\omega$ to make source samples $S_X$ look like the target domain samples $T_X$ for $\mathcal{H}_\omega$ and get a new domain $A_X$. In the third step, we pretrain the source classifier $\mathcal{F}_\theta$ on the resulted domain $A_X$, instead of the real source $S_X$. Due to the adversarial attacks' transferability, $A_X$ adversarial examples are transferable to the $\mathcal{F}_\theta$ classifier too, and actually, increase $\mathcal{F}_\theta$ accuracy on the target domain. Finally, we apply domain adaptation for the source classifier $\mathcal{F}_\theta$, using the arbitrary distance-based domain adaptation techniques~\cite{GaninL15,cdan, DAN, JAN}, but also replacing the real source data $S_X$ with the $A_X$ in the training pipeline. As a result, we obtain a multi-step domain adaptation pipeline~\cite{DAsurvey}, where at the first step, the algorithm creates a new dataset that is used by the second step method. 

\section{Experiments}
\textbf{Datasets:} All experiments were conducted in unsupervised settings (only source labels are known) on the range of digits domains (MNIST~\cite{mnisthandwrittendigit}, USPS ~\cite{uspsdataset}, MNIST-M, SVHN) and  Office-Home~\cite{venkateswara2017deep} domains.

\textbf{Settings:} In our experiments, we used $A_X$ instead of the real source $S_X$ for prominent adversarial-based approach's DANN~\cite{GaninL15}, CDAN, CDAN-E~\cite{cdan}. Also, we tested our method on Maximum Mean Discrepancy (MMD)~\cite{mmd} based on domain adaptation techniques like DAN~\cite{DAN} and JAN~\cite{JAN}. In addition, we tested our method in connection to the SHOT~\cite{SHOT} method on the Office-Home domains. We used ADA~\cite{adalib2020} library for training and testing, and all training settings were set equal to the default parameters proposed by ADA.  

The $\mathcal{H}_\omega$ domains classifier was trained 5 epochs using Adam optimizer \cite{KingmaB14} with a $0.01$ learning rate.  For $A_X$ dataset generation $l_inf$ FSGD attack with perturbation size $\varepsilon$ equal to $0.01$ and 7 steps was used. 

Before applying domain adaptation, we pretrained the $\mathcal{F}_\theta$ classifier by 10 epochs on the $A_X$ dataset. Then, each domain adaptation method used 20 epochs of training using the target domain samples. For parameters updating during the domain adaptation, we used the Adam optimizer with a learning rate equal to $0.01$. All hyperparameters were equal for each tested domain adaptation method and each pair of source and target domains.

\textbf{Results:} Averaged results over three random seeds for the best-selected hyperparameters for each adaptation task are presented in Table~\ref{tab:table-1},\ref{tab:table-2}. Across all benchmarks, the training with a $A_x$-adversarial domain makes reasonable gains over the basic settings. We find that different domains require different sizes of perturbations to generate transferable examples. Searching adversarial attack parameters more carefully individually for each task can increase adaptation accuracy.

\section{Conclusion and future works}
Our paper is still a work in progress, but we propose a simple remedy to improve the accuracy of domain adaptation methods. The proposed method allows minimizing the $\mathcal{H} \Delta \mathcal{H}$-distance between the source and target domains more successfully. By combining our method with domain adaptation techniques, we hope it may probably result in the development of less complicated and more efficient domain adaptation techniques. 
In the future, we plan to test our method on other models and other datasets, and also we are going to increase the $\mathcal{H} \Delta \mathcal{H}$-divergence attack features transferability via linearization of backpropagation~\cite{linearization} and adversarial robustness training for robust features transfer. In addition, we are planning to test our method with more advanced techniques for adversarial attacks.


\newpage

\bibliography{example_paper}
\bibliographystyle{icml2022}

\end{document}